\documentclass[preprint,12pt]{elsarticle}

\usepackage{amssymb}
\usepackage{kotex}          
\setcitestyle{square}   

\usepackage[margin=2.5cm]{geometry}

\usepackage{algorithm2e} 
\usepackage{subcaption} 
\usepackage{amsfonts} 
\usepackage{amsmath} 
\usepackage{bbm}
\usepackage[flushleft]{threeparttable}
\usepackage{caption}
\usepackage{multirow}
\usepackage{booktabs}
\usepackage{colortbl}
\usepackage{hhline,makecell}

\usepackage{float}
\usepackage{setspace}

\PassOptionsToPackage{hyphens}{url}\usepackage{hyperref}

\journal{Decision Support Systems}

\begin{document}

\begin{frontmatter}

\title{Automatic Construction of Context-Aware Sentiment Lexicon in the Financial Domain Using Direction-Dependent Words}

\author[snu]{Jihye~Park}
\ead{jihyeparkk@dm.snu.ac.kr}
\author[snu]{Hye Jin~Lee}
\ead{hyejinlee@dm.snu.ac.kr}
\author[snu]{Sungzoon~Cho\corref{cor1}}
\ead{zoon@snu.ac.kr}
\cortext[cor1]{Corresponding author}

\address[snu]{Department of Industrial Engineering \& Institute for Industrial Systems Innovation, Seoul National University, 1 Gwanak-ro, Gwanak-gu, Seoul 08826, Republic of Korea}

\begin{abstract}
Increasing attention has been drawn to the sentiment analysis of financial documents. The most popular examples of such documents include analyst reports and economic news, the analysis of which is frequently used to capture the trends in market sentiments. On the other hand, the significance of the role sentiment analysis plays in the financial domain has given rise to the efforts to construct a financial domain-specific sentiment lexicon. Sentiment lexicons lend a hand for solving various text mining tasks, such as unsupervised classification of text data, while alleviating the arduous human labor required for manual labeling. One of the challenges in the construction of an effective sentiment lexicon is that the semantic orientation of a word may change depending on the context in which it appears. For instance, the word ``profit" usually conveys positive sentiments; however, when the word is juxtaposed with another word ``decrease," the sentiment associated with the phrase ``profit decreases" now becomes negative. Hence, the sentiment of a given word may shift as one begins to consider the context surrounding the word. In this paper, we address this issue by incorporating context when building sentiment lexicon from a given corpus. Specifically, we construct a lexicon named Senti-DD for the Sentiment lexicon composed of Direction-Dependent words, which expresses each term a pair of a directional word and a direction-dependent word. Experiment results show that higher classification performance is achieved with Senti-DD, proving the effectiveness of our method for automatically constructing a context-aware sentiment lexicon in the financial domain.
\end{abstract}


\end{frontmatter}


\section{Introduction}
\label{sec:intro}

Sentiment analysis\footnote{In literature, sentiment analysis extends beyond polarity detection to include a wider range of topics. In this paper, however, we construe sentiment analysis as the task of polarity detection and use these terms interchangeably.} aims to detect sentiment polarity of a sentence, paragraph, or a document, based on its textual contents \citep{malo2014good}. It has attracted great attention not only from academia but also from the industry due to its applicability to a wide range of target populations, including consumers, companies, banks, or the general public \citep{ruiz2012semantic}. Sentiment lexicons have been the popular choice of the vehicle for conducting sentiment analysis \citep{sohangir2018financial, bonta2019comprehensive, choi2020less}. A sentiment lexicon is a list of words or phrases mapped to positive or negative sentiment labels. Such lexicons have been widely utilized across many different domains in order to solve a variety of text mining tasks, including automated sentiment computation or unsupervised document classification, while alleviating the burden of human labor required to manually label words and/or documents \citep{loughran2016textual, loughran2015use}.

In terms of sentiment analysis, the construction of a domain-specific lexicon is particularly important because the semantic orientations of words may vary by domain. This is especially true for analyzing documents in the finance domain. For instance, ``liability,” a term generally considered to be negative in common sense, is actually neutral when used in the financial domain \citep{cortis2017semeval}. Another classic example includes the terms ``bear" and ``bull." Usually, these two words do not convey any particular sentiment; yet, in a financial context, ``bear" becomes an extremely negative word (as in a ``bear market"), while ``bull" is strongly positive (as in a ``bull market"). These examples highlight the essentiality of incorporating domain-specificity when solving sentiment analysis tasks.

In the meantime, Loughran-McDonald Word List \citep{loughran2011liability} has been the popular choice of sentiment lexicon. Loughran-McDonald Word List consists of unigrams — that is, a word containing a single token. A unigram lexicon works well with words whose associated sentiment is relatively straightforward, such as ``profitable” or ``unprofitable,” with the former being obviously positive, while the latter, negative. However, a unigram-based lexicon may not successfully capture the true sentiment of a given word as its semantic orientation changes according to the context in which the word is used. That is, in other words, the sentiment of a given word may shift as one begins to consider the context which surrounds the given word. The following example illustrates such a case. The word ``profit” is usually associated with positive sentiments; however, when it is juxtaposed with another word, ``decrease,” the phrase ``profit decreases” now conveys a very negative sentiment as compared to the word ``profit” appearing by itself as a single word. In order to tackle such limitations, several attempts have been made to incorporate contextual information into the lexicon. Particularly, \cite{oliveira2016stock} have considered modifiers such as intensifiers (e.g., ``more," ``increase," ``up") and diminishers (e.g., ``less," ``decrease," ``down") to build a sentiment lexicon using microblog-messages from StockTwits. However, the consideration of the relationship between the given words and the modifiers, in this case, is still indirect, because it relies on the aggregated estimates of the degree of their association.

In this paper, we focus on context-oriented automation of a sentiment lexicon in the financial domain by extracting direction-dependent words. A word is defined to be direction-dependent if its sentiment label changes when used in combination with directional words. Direction-dependent words play a particularly important role in finance, for financial documents use these words frequently in order to depict current issues or to assess the performance of a financial entity. Consider a sentence describing the ``operating loss" of a given firm, for example. When used alone, the word ``operating loss" is usually associated with a negative sentiment, leading the sentence to be labeled negative. Yet, if juxtaposed with the directional word ``decrease," the negativity of the word ``operating loss" subsides greatly, leading the overall sentiment of the sentence to a more positive side. Similar cases can easily be observed in corporate memorandums, analyst reports, and news articles analyzing the financial market. Accurate and efficient detection of direction-dependent words from these documents will certainly help to capture the correct sentiment of financial text, producing more powerful and insightful results from the analysis. 

Previous studies \citep{malo2014good, krishnamoorthy2018sentiment} have relied on manually assembled lexicons to account for direction-dependent words when learning sentiment classifiers. It is, however, very costly to compile such lexicons manually, for it requires a great amount of background knowledge, time, and human labor during the process. Domain experts have to review one word after another in order to determine proper sentiment labels, yet there is a limit to which human efforts may cover. Furthermore, since these sentiment labels are based on the domain experts' judgment and/or intuition, they still suffer from inconsistencies due to human subjectivity.

In order to address issues associated with manual compilation and to meet the growing demand for sentiment lexicons specifically designed for financial analysis, statistical approaches \citep{mao2014automatic, oliveira2016stock} were suggested to automatically build sentiment lexicons from finance-related documents. However, these studies tended to focus only on the association between words and sentiment labels without effectively reflecting the context, and little attention has been paid to the additional challenge that the same word may represent different sentiments in different contexts.

In this paper, we propose a framework to solve the sentiment classification problem with an automatically constructed, context-aware sentiment lexicon designed specifically for use in the financial domain. The proposed framework consists of two main steps: automatic construction of the lexicon and sentiment classification. By identifying the direction-dependent words from a finance-related labeled corpus, we build our proposed sentiment lexicon and name it Senti-DD\footnote{The code is available at \url{https://github.com/sophia-jihye/Senti-DD}} for the ``Sentiment lexicon composed of Direction-Dependent words." Each element of Senti-DD is a pair consisting of a directional word and a direction-dependent word. By making use of this newly created lexical resource, we refine the sentiment score to reflect the context when solving the sentiment classification problem. The effectiveness of our method is investigated through experiments using the Financial Phrase Bank dataset, which is a resource containing around 5,000 sentences from English news headlines in the financial domain. The proposed framework using the augmented lexicon with Senti-DD achieves higher sentiment classification performance than without Senti-DD.

Our work contributes to the existing literature in three ways. First, we propose a data-driven method to automatically construct a context-aware lexicon named Senti-DD. Second, we show that the automatically identified direction-dependent words are useful for analyzing sentiments of financial documents. Third, we achieve higher sentiment classification performance by proposing an overall framework that integrates Senti-DD as a plug-in lexicon to an existing lexicon such as Loughran-McDonald Word List.

The rest of this paper is organized as follows. Section~\ref{sec:related} summarizes past literature on manual and automatic approaches to construct sentiment lexicons in the financial domain. Section~\ref{sec:proposed} describes the proposed framework in detail. Information on the dataset used in the experiment and experimental design details are reported in Section~\ref{sec:experiment}. Section~\ref{sec:results} presents and discusses the experiment results. Finally, Section~\ref{sec:conclusion} concludes our work and discusses future work.

\section{Related Works}
\label{sec:related}

In recent years, there has been increasing attention on the sentiment analysis of financial documents \citep{malo2014good}. Much of the past research has focused on the construction of a sentiment lexicon covering a wide range of domains, a major branch of which includes finance. Among those, the Harvard General Inquirer word-list has been one of the most popular lexicons used by a number of financial researchers in the earlier studies of sentiment computation from given texts \citep{stone1962general}. However, McDonald et al. \cite{loughran2011liability} has claimed that almost three-fourths (73.8\%) of the negative word found in the Harvard General Inquirer word-list are actually associated with non-negative sentiments when looked at from the perspective of business applications. After examining the quality of word classifications given by the Harvard General Inquirer word-list in a financial context, McDonald et al. \cite{loughran2011liability} proposed a revised word-list, called Loughran-McDonald Word List \cite{loughran2011liability}, which is suggested to be a better fit for financial texts. Loughran-McDonald Word List has been widely used in many financial applications, a major example of which includes tone analysis of financial news articles \citep{dougal2012journalists, garcia2013sentiment, ghosal2017iitp, gurun2012don, liu2013role, nasim2017iba}. However, as described earlier, Loughran-McDonald Word List does not reflect the context. It does not include words such as ``profit” or ``liability,” whose semantic orientation changes to a positive or negative sentiment when used with a word indicating a direction such as ``increase” or ``decrease.” As part of an attempt to build a lexicon that considers the context, Malo et al. \cite{malo2014good} constructed a list of financial entities affecting the sentiment of a sentence when used with motion verbs. After extracting a collection of financial entities from the Investopedia website, the authors pruned the collection down to 684 entities which were considered to appear most frequently in financial documents. Via manual inspection, the authors concluded that 177 out of these 684 entities affect sentiment when combined with motion verbs. However, their lexicon is manually constructed, which inevitably requires intensive human labor and time \citep{wu2019automatic}. In order to address such issues, we propose to automate the process of financial lexicon construction by incorporating pointwise mutual information.

To date, several attempts have been made to automate the extraction of lexical items from a given financial corpus by exploiting various statistical measures \citep{beigi2021automatic, moreno2020design, wu2016towards, huang2014automatic}. Oliveira et al. \cite{oliveira2016stock} utilized modifiers to incorporate contextual information into the lexicon. A modifier is an optional element in phrase or clause structure that modifies the meaning of another element in the structure. The authors measured scores for the degree of the association not only with the sentiment labels (bullish or bearish) but also with the following types of modifiers: intensifiers (e.g., ``more," ``increase," ``up") and diminishers (e.g., ``less," ``decrease," ``down"). After multiplying the two scores for each word, the authors extracted high-scoring words as words to be added to the lexicon. Their proposed method showed strong performance in sentiment classification against the six most popular lexical resources such as Harvard General Inquirer \citep{stone1962general}, MPQA subjectivity lexicon \citep{wilson2005opinionfinder}, and SentiWordNet \citep{baccianella2010sentiwordnet}. However, the consideration of the relationship between the given words and the modifiers, in this case, is still indirect, because it relies on the aggregated estimates of the degree of their association. If the words highly associated with the modifiers can be directly identified, then they can be explicitly accounted for during the lexicon construction process. Table~\ref{tab:examples} lists examples of positive and negative words found in the lexicons of previous studies.

\begin{table}
\centering
\caption{Examples of words in the lexicons suggested in the previous studies}
\label{tab:examples}
\arrayrulecolor{black}
\begin{tabular}{llll} 
\toprule
                                & \begin{tabular}[c]{@{}l@{}}Construction\\Approach\end{tabular} & Positive               & Negative           \\ 
\midrule
Harvard IV word lists [22]      & Manual                                                         & profit              & tax, liability  \\
Loughran-McDonald Word List [7] & Manual                                                         & profitable          & unprofitable    \\
Malo et al. [1]                 & Manual                                                         & (profit, up)        & (profit, down)  \\
Oliveira et al. [10]            & Automatic                                                      & profit, more profit & less profit   \\
Senti-DD (proposed)             & Automatic                                                      & (profit, up)        & (profit, down)  \\
\bottomrule
\end{tabular}
\arrayrulecolor{black}
\end{table}

In this paper, we employ directional words to tackle the limitations of the past arising from the manual compilation and/or the inefficient/indirect consideration of the contextual information. Directional words can be regarded as one of the important word-groups among modifiers because they affect the sentiment of direction-dependent words. Therefore, it is important to consider the co-occurrence of directional words and direction-dependent words in conducting sentiment analysis. We propose to automate the process of constructing a financial lexicon (1) by directly extracting direction-dependent words based on the measure of association between the given word and the direction-dependency type and (2) by expanding the lexicon by adding positive/negative-context pairs consisting of directional words and direction-dependent words.

\section{Proposed Approach}
\label{sec:proposed}

\begin{figure}
\center
\includegraphics[width=0.99\textwidth]{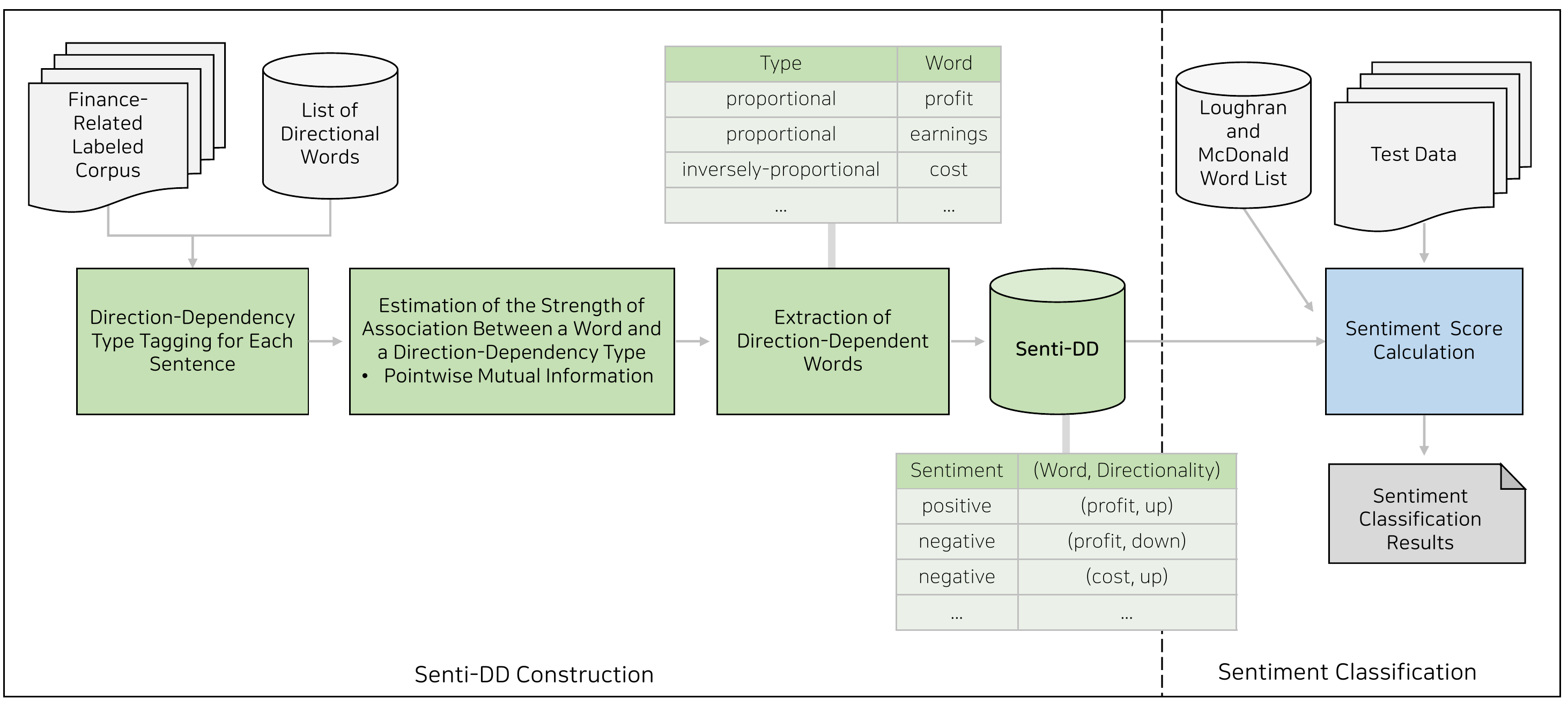} 
\caption{An overview of the proposed sentiment analysis framework}
\label{fig:overview}
\end{figure}

In this section, we introduce a data-driven sentiment analysis framework, which comprises two stages: (1) Senti-DD construction, and; (2) sentiment classification. The first stage builds Senti-DD computing pointwise mutual information (PMI) score, based on which a given word's direction-dependency type is estimated. The second stage, then, solves the sentiment classification problem using the Senti-DD built during the first stage. An overview of the proposed framework is illustrated in Fig.~\ref{fig:overview}.

\subsection{Senti-DD Construction}

\subsubsection{Direction-dependency Type Tagging for Each Sentence}
\label{sec:tagging}

First, we gather polar sentences from a finance-related labeled corpus. A polar sentence is a sentence representing either a positive or negative sentiment. Given a set of polar sentences and a word list with each word assigned to a directional label ``up" or ``down," we define the $UpScore$ for each sentence $s$ as the number of ``up" words found in the subject sentence. The $DownScore$, similarly, is defined as the number of ``down" words found in the subject sentence. Finally, we compute the direction score of a given sentence as follows:
\begin{equation}
DirectionScore(s) = UpScore(s) - DownScore(s)\,,
\end{equation}
This direction score reflects the degree of direction the given sentence conveys. In this paper, we rely on the Financial Phrase Bank \citep{malo2014good} and the Harvard General Inquirer word-list \citep{stone1962general} as the choice for the finance-related labeled corpus and direction-labeled word list, respectively. We present the details on the summary statistics of the datasets in Section~\ref{sec:dataset}.

Finally, considering the relationship between a direction score and a sentiment label of a sentence, we give each sentence a tag representing a direction-dependency type. The direction-dependency tags are categorized into two types: ``proportional" and ``inversely proportional." A sentence is tagged as a ``proportional" type if its sentiment is either positive with a direction score greater than zero or negative with a direction score less than zero. Similarly, a sentence is tagged as an ``inversely proportional" type if its sentiment is either positive with a direction score less than zero or negative with a direction score greater than zero. 

\subsubsection{Estimation of the Strength of Association Between a Word and a Direction-Dependency Type}
\label{sec:estimation}

Prior to estimating statistical correlations, we perform several preprocessing steps on the sentences with the tags. We transform each sentence into a list of nouns by performing tokenization, part-of-speech tagging, and lemmatization using the NLTK\footnote{\url{http://www.nltk.org}} library. We have only kept words that appeared more than five times in the entire corpus to discard less frequent words. Furthermore, we have also removed words the length of which is one since those words are often the cases that are difficult to interpret.

We propose to measure the association between a word $w$ and its direction-dependency type,  which is either a ``proportional" type $t_p$ or an ``inversely proportional" type $t_i$, by exploiting the definition of the PMI score. More specifically, we define the direction-dependency score of a given word as follows:
\begin{equation}
  DirectionDependencyScore(w) =
    \begin{cases}
      |PMI(w,t_p)| & \text{if $PMI(w,t_p) > PMI(w,t_i)$}\\
      0 & \text{if $PMI(w,t_p) = PMI(w,t_i)$}\\
      -|PMI(w,t_i)| & \text{if $PMI(w,t_p) < PMI(w,t_i)$}
    \end{cases}
\end{equation}
\begin{equation}
  , \text{where } PMI(w,t)=log_2{\frac{p(w,t)}{p(w)p(t)}} \,,
\end{equation}
where $p(w,t)$ is the probability that the sentence of direction-dependency type $t$, containing the word $w$, is found in the subject corpus; $p(w)$ is the probability that the word $w$ is found in the subject corpus; and $p(t)$ is the probability that the sentence of direction-dependency type $t$ is found in the subject corpus. 

A word with a positive direction-dependency score is regarded as a candidate word of a ``proportional" type, which would represent a positive sentiment when used with ``up" words and represent a negative sentiment when used with ``down" words. Similarly, a word with a negative direction-dependency score is regarded as a candidate word of an ``inversely proportional" type, which would represent a positive sentiment when used with ``down" words and represent a negative sentiment when used with ``up" words. 

\subsubsection{Extraction of Direction-dependent Words}

Based on the relationship between the direction-dependency tag of a sentence and the direction-dependency score of a word computed in Section~\ref{sec:estimation}, we extract a single representative word for direction-dependency from each sentence according to the following rules: if a sentence is a ``proportional" type, the word whose direction-dependency score is the highest among the candidates of the ``proportional" words is extracted as a ``proportional" type direction-dependent word; and if a sentence is an ``inversely proportional" type, the word whose direction-dependency score is the lowest among the candidates of the ``inversely proportional" words is extracted as an ``inversely proportional" type direction-dependent word. 

\subsubsection{Senti-DD Construction based on the List of Directional Words and the Direction-dependent Words}

To construct Senti-DD, we create pairs of words from the list of directional words and the list of direction-dependent words. A pair of an ``up" word and a ``proportional" word, or that of a ``down" and an ``inversely proportional" word, is labeled as a positive-context pair. Similarly, a pair of an ``up" word and an ``inversely proportional" word or that of a ``down" word and a ``proportional" word is labeled as a negative-context pair. 

\subsection{Sentiment Classification}

The classification is performed based on the augmented lexicon combining Loughran-McDonald Word List and Senti-DD. Based on Loughran-McDonald Word List, we first determine the overall polarity by calculating the sentiment score for each sentence. The score is then refined based on the proposed Senti-DD to capture the occurrence of the direction-dependent words as well as directional words. Finally, based on the refined score, the sentence is then classified as a positive, negative, or neutral class.

\subsubsection{Sentiment Score Calculation based on Loughran-McDonald Word List}
\label{sec:lm}

First of all, we calculate the sentiment score for each sentence based on Loughran-McDonald Word List. We define the $PosScore$ for a sentence $s$ as the number of positive words found in the subject sentence. The $NegScore$, similarly, is defined as the number of negative words found in the subject sentence. Finally, we compute the sentiment score of a given sentence as follows:
\begin{equation}
\label{eq:lm}
SentimentScore(s) = PosScore(s) - NegScore(s) \,,
\end{equation}

\subsubsection{Sentiment Score Refinement based on Senti-DD}

To refine the sentiment score, we calculate a sentiment score reflecting context. Using Senti-DD, we define the $ContextPosScore$ for a sentence $s$ as the number of positive-context pairs found in the subject sentence. The $ContextNegScore$, similarly, is defined as the number of negative-context pairs found in the subject sentence. We then compute the sentiment score reflecting the context as follows: 
\begin{equation}
ContextSentimentScore(s) = ContextPosScore(s) – ContextNegScore(s) \,,
\end{equation}
Finally, to reflect the extra-positivity/negativity driven by the context of the sentence, we refine the sentiment score based on the $ContextSentimentScore$. We add one point to the sentiment score if the $ContextSentimentScore$ is greater than zero. Meanwhile, we subtract one point from the sentiment score if the $ContextSentimentScore$ is less than zero. As a result, we obtain the refined sentiment score as follows: 
\begin{equation}
  RefinedSentimentScore(s) =
    \begin{cases}
      SentimentScore(s)+1 & \text{if $ContextSentimentScore > 0$}\\
      SentimentScore(s) & \text{if $ContextSentimentScore = 0$}\\
      SentimentScore(s)-1 & \text{if $ContextSentimentScore < 0$}
    \end{cases}       
\end{equation} 

\subsubsection{Sentiment Classification based on the Refined Sentiment Score}

The sentence is then classified according to the following rules: if the refined sentiment score is greater than zero, the sentence is classified as a positive class; if the score is less than zero, the sentence is classified as a negative class; and if the score equals to zero, the sentence is classified as a neutral class.

\section{Experiment}
\label{sec:experiment}

\subsection{Dataset Description}
\label{sec:dataset}

In our experiment, we use the Financial Phrase Bank\footnote{\url{https://www.researchgate.net/publication/251231364_FinancialPhraseBank-v10}} dataset created by Malo et al \cite{malo2014good}. The Financial Phrase Bank is a labeled corpus of financial news headlines. The dataset consists of 4,835 English sentences, which were annotated by 16 people with a background in finance and business. The annotators were asked to give a positive, negative, or neutral label according to how they think the information in the sentence might affect the stock price of the mentioned company. Based on the level of agreement among the annotators, the dataset is divided into four subsets: DS50 (more than 50\% agreement), DS66 (more than 66\% agreement), DS75 (more than 75\% agreement), and DS100 (100\% agreement among annotators). The total number of sentences and distribution of labels on each dataset is given in Table~\ref{tab:dataset}.

To ensure the robustness of the results, we make use of the stratified 5-fold cross-validation. In stratified 5-fold cross-validation, the original dataset is partitioned into five equal-sized folds, with each fold is an appropriate representative of the original data with respect to the label distribution. Of the five folds, a single fold is retained as the test data, and the remaining four folds are used as training data. The cross-validation process is repeated five times, with each of the five folds used exactly once as the test data. We then report an average over the five results. 

\begin{table}
\centering
\caption{Dataset characteristics}
\label{tab:dataset}
\arrayrulecolor{black}
\begin{tabular}{ccccc} 
\toprule
Dataset & \%Positive & \%Neutral & \%Negative & Number of sentences  \\ 
\midrule
DS50   & 28.2       & 59.3      & 12.5       & 4,835                \\
DS66    & 27.8       & 60.0      & 12.2       & 4,209                \\
DS75    & 25.7       & 62.1      & 12.2       & 3,447                \\
DS100    & 25.2       & 61.4      & 13.4       & 2,259                \\
\bottomrule
\end{tabular}
\arrayrulecolor{black}
\end{table}

\subsection{Details on Building Senti-DD}
\label{sec:details}

To define the directional words used in Section~\ref{sec:tagging}, we follow the experimental settings of previous works \citep{malo2014good, krishnamoorthy2018sentiment}: we form two word-lists, a list for ``up" terms and a list for ``down" terms, by using the Harvard General Inquirer word-list \citep{stone1962general} as seed lists. Words defined in the Harvard word lists under categories of ``increase” and ``rise” are classified as ``up" terms, and the words under categories of ``decrease” and ``fall” are classified as ``down" terms. By manually reviewing these words, we not only exclude words that have different meanings in the financial context but also include several additional words including ``award” that describe a change of an event \citep{krishnamoorthy2018sentiment}. The final number of words in the list is 31, with 20 terms for the ``up" category and 11 terms for the ``down" category. Table~\ref{tab:directional} shows the full list of carefully selected directional words. After stemming both the words in each sentence of the Financial Phrase Bank and the directional words, we compare them for matches. 

\begin{table}
\centering
\caption{List of directional words}
\label{tab:directional}
\arrayrulecolor{black}
\begin{tabular}{cl} 
\toprule
Directionality type & \multicolumn{1}{c}{Words}\\ 
\midrule
Up   & \begin{tabular}[c]{@{}l@{}} accelerate, advance, award, better, climb,\\double, faster, gain, grow, higher,  increase,\\jump, quicken, rebound, recover, rise, rose,\\step-up, surge, up \end{tabular}  \\ 
\hline
Down & \begin{tabular}[c]{@{}l@{}} constrain, decelerate, decline, decrease,\\down,  drop, fall, fell, slower, weaken, weaker \end{tabular}                                                                    \\
\bottomrule
\end{tabular}
\arrayrulecolor{black}
\end{table}

In this experiment, 675 sentences are tagged as a ``proportional" type and 27 sentences are tagged as an ``inversely proportional" type, where the total number of sentences for the dataset used in the experiment is 4,835. Table~\ref{tab:sentences} illustrates examples of the sentences with the tags.

\begin{table}
\centering
\caption{Examples of the sentences tagged as a ``proportional" or ``inversely proportional" type}
\label{tab:sentences}
\resizebox{\columnwidth}{!}{%
\begin{tabular}{cl} 
\toprule
Direction-dependency type & \multicolumn{1}{c}{Sentences}  \\ 
\midrule
Proportional              & \begin{tabular}[c]{@{}l@{}} - In the fourth quarter of 2009, Orion's net profit went up by 33.8\% year-on-year to EUR 33m.\\- Agricultural newspaper Maaseudun Tulevaisuus had 318,000 readers, representing a decrease of 6\%. \end{tabular}  \\ 
\hline
Inversely proportional    & \begin{tabular}[c]{@{}l@{}} - Unit costs for flight operations fell by 6.4 percent.\\- Operating loss increased to EUR 17 mn from a loss of EUR 10.8 mn in 2005. \end{tabular}                         \\
\bottomrule
\end{tabular}
}
\end{table}

\subsection{Evaluation Metrics}
We employ the following three metrics to evaluate the performance of our proposed model:
\begin{itemize}
	\item Precision is the proportion of true positives among the cases predicted as positive. It measures the correctness of the model. Mathematically it is defined as:
	\begin{equation}
    Precision = \frac{TP}{TP+FP},
    \end{equation}

	\item Recall is the proportion of true positives among the positive cases. It measures the completeness of the classification result. Mathematically it is defined as:
	\begin{equation}
    Recall = \frac{TP}{TP+FN},
    \end{equation}

	\item F1 score is a harmonic mean of precision and recall. It measures the general accuracy of the model prediction. Mathematically it is defined as:
	\begin{equation}
    F1 \text{ } score = 2 \times \frac{Precision \times Recall}{Precision + Recall},
    \end{equation}
\end{itemize}

\subsection{Baseline Lexicons}
\begin{itemize}
	\item SentiWordNet (SWN): SentiWordNet \citep{baccianella2010sentiwordnet} is a lexical resource explicitly devised for supporting sentiment classification and opinion mining applications. SentiWordNet uses synsets (sets of synonyms) for 117,659 terms, each of which is associated with three numerical scores describing how objective, positive, and negative the terms contained in the synsets are. A sentiment score for each sentence is obtained by subtracting the sum of negative scores from the sum of positive scores for words in the sentence. In our experiment, we classified a sentence as a positive class if the sentiment score was greater than zero, a negative class if the score was less than zero, and a neutral class if the score was zero. 
	\item TextBlob: TextBlob\footnote{\url{https://github.com/sloria/TextBlob}} \citep{loria2018textblob} is a widely distributed library for sentiment analysis. TextBlob uses the Pattern \citep{de2012pattern} module, which contains a lexicon\footnote{\url{https://github.com/sloria/TextBlob/blob/dev/textblob/en/en-sentiment.xml}} for 2,918 English adjectives and adverbs, each of which is mapped to polarity, subjectivity, and intensity scores. Using the lexicon, TextBlob returns a polarity score ranging from -1 (most negative) to 1 (most positive). In our experiment, we classified a sentence as a positive class if the polarity score was greater than zero, a negative class if the score was less than zero, and a neutral class if the score was zero.
	\item VADER: VADER \citep{hutto2014vader} is a
    general-purpose sentiment analysis library based on a curated lexicon of 7,517 words. VADER has been quite successful when dealing with social media texts, movie reviews, and product reviews. VADER calculates a compound sentiment score by weighting each word in a given text according to several rules that change the intensity of the word. VADER returns a normalized compound score ranging from -1 (most negative) to 1 (most positive). Consistent with prior applications of the VADER library, a compound score greater than 0.05 is classified as a positive class, the score less than -0.05 is classified as a negative class, and the score between -0.05 and 0.05 is classified as a neutral class.
	\item Loughran-McDonald Word List (LM): Loughran-McDonald Word List \citep{loughran2011liability} is one of the most popular sentiment lexicons in the financial domain. This resource contains 354 positive words and 2,355 negative words. We used the sentiment score defined by Equation~\ref{eq:lm} in Section~\ref{sec:lm}.
\end{itemize}

\section{Results and Discussion}
\label{sec:results}

\subsection{Quantitative Evaluation}

Table~\ref{tab:classification} reports how each lexicon performs on the sentiment classification task as compared to the augmented lexicon with Sent-DD. SWN, TextBlob, VADER, and LM stand for the sentiment classification method based on SentiWordNet, TextBlob, VADER, and Loughran-McDonald Word List, respectively; and LM+Senti-DD stands for the proposed framework using the augmented lexicon combining Loughran-McDonald Word List and Senti-DD. All values are weighted average values and best values per measure are marked in bold. The results indicate that LM+Senti-DD always outperforms other baseline lexicons by consistently achieving higher F1 scores. Given that the level of agreement among the annotators of the four datasets used in the experiment varies and that a low level of agreement means low-quality labels, the proposed method shows robust performance against variations in labeling quality. This indicates that the score refinement using Senti-DD, which reflects the context by capturing the co-occurrence of directional words and direction-dependent words, is effective not only for high-quality but also for low-quality labels.

\begin{figure}
\center
\includegraphics[width=0.99\textwidth]{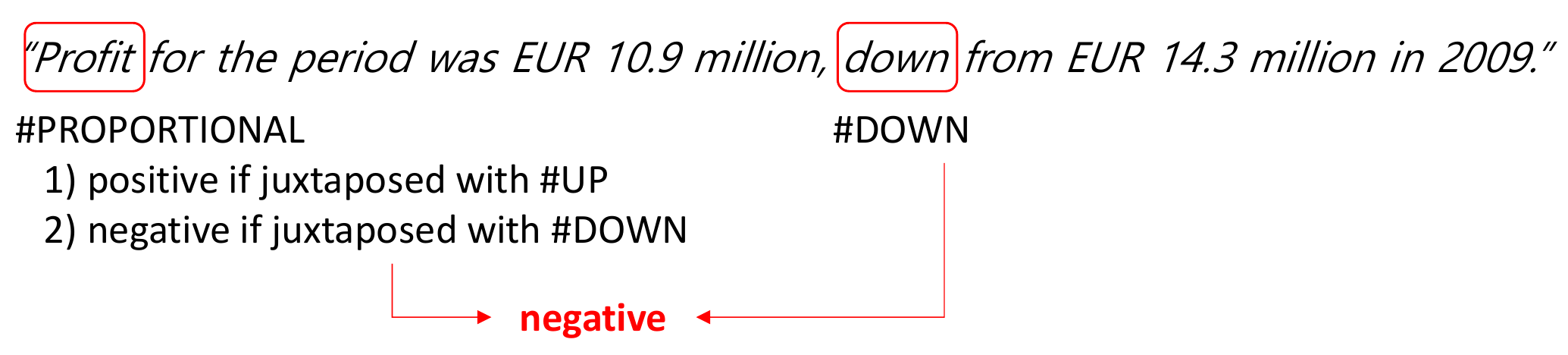} 
\caption{Example of performing sentiment analysis using Senti-DD}
\label{fig:example}
\end{figure}

\begin{figure}
    \centering
    
    \begin{subfigure}[t]{\textwidth}
        \centering
        \includegraphics[width=0.99\textwidth]{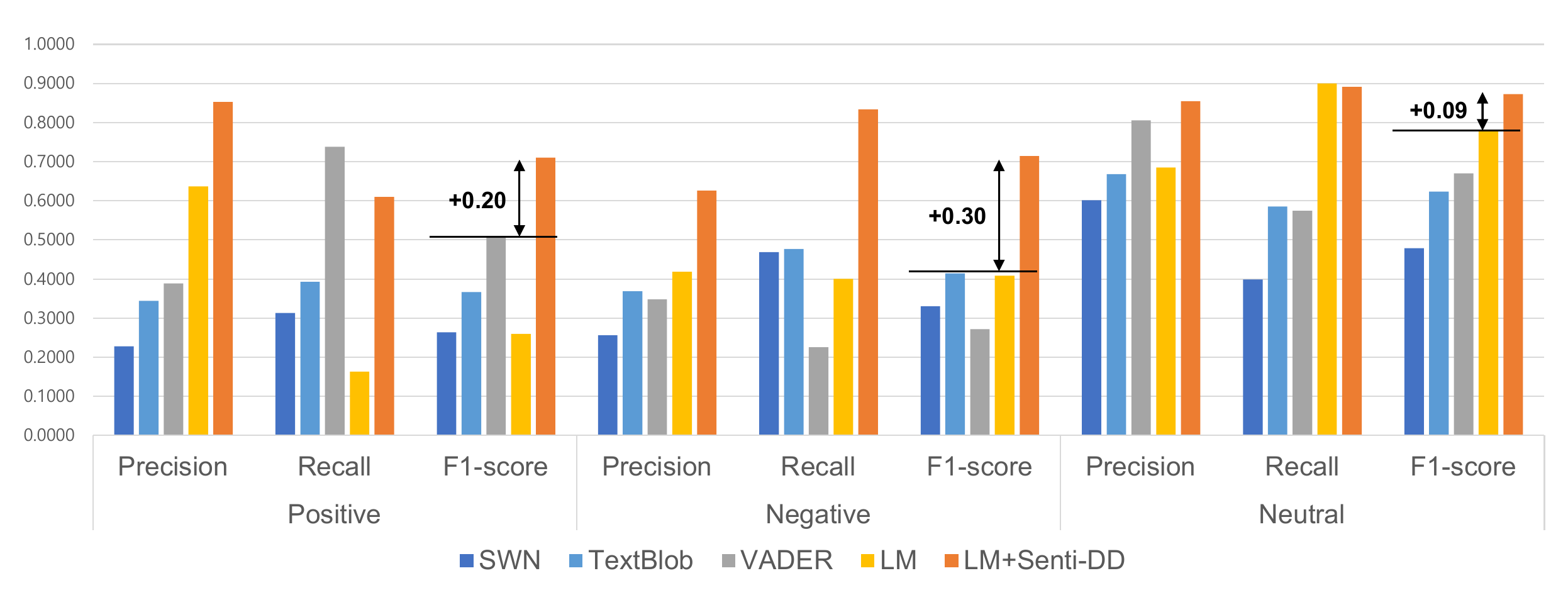} 
        \caption{Graphical comparison of the results for the DS100 dataset}
        \label{fig:graphical}
    \end{subfigure}
    
    \vspace{1cm}
    
    \begin{subtable}[t]{\textwidth}
    \centering
    \caption{Experiment results of the classification task on the DS100 dataset}
    \label{tab:classification-DS100}
    \resizebox{\columnwidth}{!}{%
    \arrayrulecolor{black}
    \begin{tabular}{cc!{\vrule width \lightrulewidth}cccccc} 
\toprule
Dataset                & Class                     & Measure   & SWN    & TextBlob & VADER  & LM              & LM+Senti-DD         \\ 
\midrule
\multirow{9}{*}{DS100} & \multirow{3}{*}{Positive} & Precision & 0.2284 & 0.3439   & 0.3889 & 0.6371          & \textbf{0.8528}  \\
                       &                           & Recall    & 0.3132 & 0.3931   & \textbf{0.7384} & 0.1632          & 0.6101  \\
                       &                           & F1-score  & 0.2637 & 0.3668   & 0.5085 & 0.2596          & \textbf{0.7106}  \\ 
\cline{2-8}
                       & \multirow{3}{*}{Negative} & Precision & 0.2562 & 0.3689   & 0.3479 & 0.4184          & \textbf{0.6264}  \\
                       &                           & Recall    & 0.4690 & 0.4768   & 0.2255 & 0.4004          & \textbf{0.8340}  \\
                       &                           & F1-score  & 0.3310 & 0.4143   & 0.2720 & 0.4087          & \textbf{0.7145}  \\ 
\cline{2-8}
                       & \multirow{3}{*}{Neutral}  & Precision & 0.6017 & 0.6686   & 0.8058 & 0.6852          & \textbf{0.8544}  \\
                       &                           & Recall    & 0.3988 & 0.5850   & 0.5743 & \textbf{0.9005} & 0.8918           \\
                       &                           & F1-score  & 0.4790 & 0.6238   & 0.6705 & 0.7782          & \textbf{0.8725}  \\
\bottomrule
\end{tabular}
    }
    \arrayrulecolor{black}
    \end{subtable}
    
    \caption{Comparative evaluation results of the classification task on the Financial Phrase Bank dataset}
\end{figure}

The key idea of the proposed framework using Senti-DD is to develop financial domain-specific clues so that it can achieve extremely high performance to predict positive or negative sentiment at the cost of a relatively low recall for the neutral class. The graphical comparison of the results for the DS100 dataset is illustrated in Fig.~\ref{fig:graphical}. What stands out in the figure is that the F1 score for a negative class achieved by LM+Senti-DD is almost twice higher than the score achieved by LM. It appears that the proposed augmented lexicon contributes to increasing overall performance by correctly classifying several sentences, which used to be misclassified as a neutral class in the approach of LM, into a positive or negative class. The key factor in improving performance is that LM+Senti-DD reflects the context by incorporating the effects of directional words in classifying sentences. For example, let us consider the following sentence:
\begin{quote}
Profit for the period was EUR 10.9 million, down from EUR 14.3 million in 2009.
\end{quote}
LM misclassifies the above sentence, a negative statement from an investor perspective, as a neutral class because the sentence contains neither positive nor negative words in Loughran-McDonald Word List. The proposed method, however, correctly classifies the sentence as a negative class since it contains both a ``down" type word, ``down,” and a ``proportional" type word, ``profit.” Fig.~\ref{fig:example} illustrates this example of performing sentiment analysis using the proposed Senti-DD.

Table~\ref{tab:classification-DS100} shows the experiment results on the DS100 dataset in detail. The results indicate that LM+Senti-DD generally outperforms other lexicons, especially in classifying the positive and the negative classes. Although VADER achieves a high recall for a positive class, it records a very low precision compared to LM+Senti-DD, leading LM+Senti-DD to achieve a higher f1 score. It appears that the rules for calculating sentiment scores in VADER tend to be biased toward predicting a large number of positive sentiments. 

\begin{table}
\centering
\caption{Experiment results of the classification task on the four datasets defined based on the Financial Phrase Bank}
\label{tab:classification}
\arrayrulecolor{black}
\begin{tabular}{cc!{\vrule width \lightrulewidth}ccccc} 
\toprule
Dataset                & Measure   & SWN \citep{baccianella2010sentiwordnet}    & TextBlob \citep{loria2018textblob} & VADER \citep{hutto2014vader}  & LM \citep{loughran2011liability}     & LM+Senti-DD         \\ 
\midrule
\multirow{3}{*}{DS50}  & Precision & 0.4778 & 0.5155   & 0.6028 & 0.6147 & \textbf{0.7090}  \\
                       & Recall    & 0.3969 & 0.4852   & 0.5396 & 0.6232 & \textbf{0.7055}  \\
                       & F1-score  & 0.4107 & 0.4953   & 0.5452 & 0.5914 & \textbf{0.7001}  \\ 
\hline
\multirow{3}{*}{DS66}  & Precision & 0.4851 & 0.5275   & 0.6194 & 0.6337 & \textbf{0.7389}  \\
                       & Recall    & 0.4044 & 0.4968   & 0.5534 & 0.6363 & \textbf{0.7315}  \\
                       & F1-score  & 0.4194 & 0.5070   & 0.5599 & 0.6023 & \textbf{0.7271}  \\ 
\hline
\multirow{3}{*}{DS75}  & Precision & 0.4916 & 0.5426   & 0.6409 & 0.6507 & \textbf{0.7796}  \\
                       & Recall    & 0.4009 & 0.5039   & 0.5590 & 0.6556 & \textbf{0.7702}  \\
                       & F1-score  & 0.4199 & 0.5169   & 0.5702 & 0.6174 & \textbf{0.7673}  \\ 
\hline
\multirow{3}{*}{DS100} & Precision & 0.4624 & 0.5476   & 0.6405 & 0.6377 & \textbf{0.8238}  \\
                       & Recall    & 0.3873 & 0.5228   & 0.5688 & 0.6476 & \textbf{0.8128}  \\
                       & F1-score  & 0.4058 & 0.5317   & 0.5770 & 0.5982 & \textbf{0.8105}  \\
\bottomrule
\end{tabular}
\arrayrulecolor{black}
\end{table}

\subsection{Qualitative Evaluation}

Table~\ref{tab:entities} illustrates all of the direction-dependent words extracted from the training process using DS50 dataset. In the table, 72 ``proportional" type words and seven ``inversely proportional" type words are listed in the ascending alphabetical order. As described in Section~\ref{sec:proposed}, ``proportional" type words may either lead to a positive sentiment when combined with an ``up" type word or lead to a negative sentiment when combined with a ``down" type word; whereas ``inversely proportional" type words may either lead to a positive sentiment when combined with a ``down" type word or lead to a negative sentiment when combined with an ``up" type word. A majority of the words seem to be appropriately identified. It is quite intuitive that ``capital,” ``demand," ``ebit” which means earnings before interest and taxes (EBIT), ``growth,” ``investment,” ``margin,” ``profit,” and ``revenue” are identified as a ``proportional" type; whereas ``cost” is identified as an ``inversely proportional" type.

\begin{table}
\centering
\caption{List of direction-dependent words extracted from the training process for the entire DS50 dataset}
\label{tab:entities}
\arrayrulecolor{black}
\begin{tabular}{cl} 
\toprule
Direction-dependency type & \multicolumn{1}{c}{Words}                                                                                                                                                                                                                                                                                                                                                                                                                                                                                                                                                                                                                                                \\ 
\midrule
Proportional              & \begin{tabular}[c]{@{}l@{}} acquisition, agreement, area, beer, brewery, business, \\capital, cash, cent, communication, contract, currency, \\customer, demand, division, ebit, efficiency, electronics, \\end, eur, finnair, food, group, growth, income,~interest, \\investment, item, june, konecranes, liter, maker, management,\\manufacturer, march, margin, medium, metal, mln, month, \\net, news, order, orion, oyj, paper, passenger, percent, period, \\phone, product, profit, property, pulp, quarter, report, revenue, \\sale, september, share, solution, system, teleste, time, tonne, \\trade, turnover, use, value, volume, world, year\end{tabular}  \\ 
\hline
Inversely proportional    & day, cost, construction, result, company, plant, traffic                                                                                                                                                                                                                                                                                                                                                                                                                                                                                                                                                                                                                 \\
\bottomrule
\end{tabular}
\arrayrulecolor{black}
\end{table}

As indicated in Section~\ref{sec:details}, it should be noted that, in the dataset used in the experiment, the number of sentences tagged as an ``inversely proportional" type is relatively small, thus leading to the ``inversely proportional" type words in a small number. Furthermore, the imbalance between the number of ``proportional" and ``inversely proportional" type sentences seems to give rise to noisy words such as ``beer” and ``day” which are not intuitively interpreted as direction-dependent words. Due to the imbalance, some words which should be frequently used regardless of direction may not appear in the ``inversely proportional" type sentences but only appear in the ``proportional" type sentences. 
Notwithstanding its limitations, these preliminary results show the possibility of automatically acquiring direction-dependent words by using the proposed PMI based method.

\section{Conclusion}
\label{sec:conclusion}

With the growing demand for sentiment analysis in financial and economic applications, it is essential to build a domain-specific sentiment lexicon in an efficient yet effective manner. 
This paper presents a framework to enrich an existing financial sentiment lexicon by combining direction-dependent words automatically extracted from a labeled corpus of financial news headlines. In the framework, Senti-DD, each element of which is a pair consisting of a directional word and a direction-dependent word, is generated by calculating PMI score to estimate the strength of association between a word and a direction-dependency type. The framework then performs sentiment classification by using Senti-DD as a plug-in lexicon to Loughran-McDonald Word List.

Experiment results show that the augmented lexicon with Senti-DD outperforms the existing financial sentiment lexicons when solving the sentiment classification task. Particularly, ``acquisition," ``agreement," and ``communication" are identified as a ``proportional" type words which lead to a positive sentiment when combined with an ``up" type word or lead to a negative sentiment when combined with a ``down" type word. These results prove that our proposed data-driven method to automatically identify direction-dependent words and to construct the context-aware lexicon Senti-DD works effectively for the sentiment classification task in the financial domain.

The proposed approach, which automatically extracts the words dependent on certain types of words, may be easily applied to other domains. It would be interesting to deal with online reviews on electronics, where semantic orientations of certain words such as ``definition” and ``price” may change depending on the context they are used. For example, ``definition” would lead to a positive sentiment when used with ``high” but lead to a negative sentiment when used with ``low.” We would like to extract these context-dependent words from various domains and investigate how the characteristics of the words differ by domain. Also, lexicon construction may always benefit from a larger amount of data. Especially, acquiring labeled datasets for financial documents with a lot of directional words, such as analyst reports, will help improve the quality of Senti-DD.

\section*{Credit authorship contribution statement}

\textbf{Jihye Park:} Conceptualization, Methodology, Investigation, Writing - original draft. \textbf{Hye Jin Lee:} Validation, Writing - original draft. \textbf{Sungzoon Cho:} Writing - review \& editing, Supervision.

\section*{Acknowledgement}

This work was supported by National Research Foundation of Korea (2018R1D1A1A02045842).


\bibliographystyle{elsarticle-num}
\bibliography{ref}

\end{document}